\documentclass{llncs}
\usepackage{llncsdoc}
\usepackage{url}
\usepackage{latexsym}
\usepackage{amsmath}
\usepackage{array}
\usepackage{multirow}
\usepackage{makecell}
\usepackage{tikz}
\usepackage{tikz-qtree}
\usepackage{epstopdf}
\usepackage[linesnumbered,ruled]{algorithm2e}

\begin{document}

\renewcommand{\arraystretch}{1.2}
\setlength{\tabcolsep}{2pt}

\title{Word Segmentation on Micro-blog Texts with External Lexicon and Heterogeneous Data} 

\author{Qingrong Xia, Zhenghua Li\thanks{Corresponding Author.}, Jiayuan Chao, Min Zhang}
\institute{Soochow University, Suzhou, China \\
\email{\{kirosummer.nlp, chaojiayuan.china\}@gmail.com, \\
 \{zhli13, minzhang\}@suda.edu.cn}}

\maketitle
\begin{abstract}
This paper describes our system designed for the NLPCC 2016 shared task on 
word segmentation on micro-blog texts (i.e., Weibo).
We treat word segmentation as a character-wise sequence labeling problem, 
and explore two directions to enhance our CRF-based baseline.
First, we employ a large-scale external lexicon for constructing extra lexicon features in the model, which is proven to be extremely useful.
Second, we exploit two heterogeneous datasets, i.e., Penn Chinese Treebank 7 (\textsl{CTB7}) and People Daily (\textsl{PD}) to help word segmentation on Weibo.
We adopt two mainstream approaches, i.e., the guide-feature based approach and the recently proposed coupled sequence labeling approach.
We combine the above techniques in different ways and obtain four well-performing models.
Finally, we merge the outputs of the four models and obtain the final results via Viterbi-based re-decoding.
On the test data of Weibo, our proposed approach outperforms the baseline by $95.63-94.24=1.39\%$ in terms of F1 score.
Our final system rank the first place among five participants in the open track in terms of F1 score, and is also the best among all 28 submissions. 
All codes, experiment configurations, and the external lexicon are released at \url{http://hlt.suda.edu.cn/~zhli}.
\end{abstract}

\section{Introduction}
\label{intro}

Chinese word segmentation (WS) is the most fundamental task in Chinese language processing.
In the past decade, supervised approaches have gained extensive progress on canonical texts, 
especially on texts from domains or genres similar to existing manually labeled data\footnote{Please refer to \url{http://zhangkaixu.github.io/bibpage/cws.html} for a long list of related papers.}.
However, the upsurge of web data imposes great challenges on existing techniques.
The performance of the state-of-the-art systems degrades dramatically on informal web texts, such as micro-blogs, product comments, and so on.
Driven by this challenge, NLPCC 2016 organizes a shared task with an aim of promoting WS on Weibo ($\textsl{WB}$, Chinese pinyin of micro-blogs) text \cite{qiu2016overview}.

This paper describes our system designed for the shared task in detail.
We treat WS as a character-wise sequence labeling problem, and build our model based on the standard conditional random field (CRF) \cite{lafferty2001conditional} with bigram features.
Our major contributions are three-fold.
First, we employ a large-scale external lexicon for constructing extra lexicon features in the model, which is proven to be extremely useful.

Second, we exploit two mainstream approaches to exploit heterogeneous data, i.e.,the guide-feature based approach and the recently proposed coupled sequence labeling approach.
The third-party heterogeneous resources used in the work are Penn Chinese Treebank 7.0 ($\textsl{CTB7}$, $50K$) and People's Daily ($\textsl{PD}$, $100K$). Since \textsl{CTB7} and \textsl{PD} have different annotation standards in word segmentation and part-of-speech (POS) tagging, \textsl{PD} has been automatically converted into the style of \textsl{CTB7}.

Third, we propose a merge-then-re-decode ensemble approach to combine the outputs of different base models. 

On the test data of Weibo, our proposed approach outperforms the baseline by $95.63-94.24=1.39\%$ in terms of F1 score.
Our final system rank the first place among five participants in the open track in terms of F1 score, and is also the best among all 28 submissions.

This paper is organized as follows. 
Section \ref{sec:crf} introduces the baseline CRF-based word segmentation model.
Section \ref{sec:lexicon} describes how to employ external lexicon features into   baseline CRF model. 
Section \ref{sec:guide} briefly illustrates the guide-feature based approach while
Section \ref{sec:coupled} briefly presents the coupled sequence labeling approach.
Section \ref{sec:ensemble} introduces the merge-then-re-decode ensemble approach.
Section \ref{sec:exp} presents the experimental results.
We discuss closely related works in Section \ref{sec:related} and conclude this paper in Section \ref{sec:conclude}.

\section{The Baseline CRF-based WSTagger}
\label{sec:crf}

We treat WS as a sequence labeling problem and employ the standard CRF with bigram features. 
We adopt the $\{B, I, E, S\}$ tag set, indicating the beginning of a word, the inside of a word, the end of a word and a single-character word \cite{xue2003chinese}.

\begin{figure}[tb]
\centering
    \includegraphics[scale=1.0]{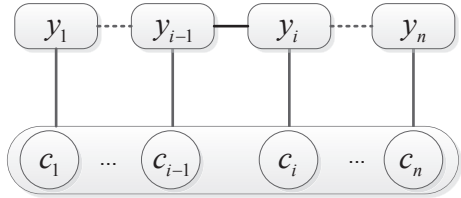}
\vspace{0.2em}
    \caption{\label{fg-tagger} Graphical structure of the baseline CRF model.}
\end{figure}

Figure \ref{fg-tagger} shows the graphical structure of the CRF model.
Given an input sentence, which is a sequence of $n$ characters, denoted by
$\mathbf{x} = c_1...c_n$,
WS aims to determine the best tag sequence $\mathbf{y} = y_1...y_n$, where $y_i \in \{B,I,E,S\}$.
As a log-linear model, CRF defines the probability of a tag sequence as:
\begin{equation}\label{eq-traditional}
	\begin{split}
  	P(\mathbf{y|x};\theta ) =& \frac{e^{Score(\mathbf{x,y};\theta )}}{\sum_{\mathbf{y}^{'}}e^{Score(\mathbf{x,y}^{'};\theta )}}   \\
  	Score(\mathbf{x,y};\theta) =& \sum_{1\leq i\leq n+1}\theta \cdot \mathbf{f}_{bs}(\mathbf{x},i,y_{i-1},y_i)
  	\end{split}
\end{equation}
where $Score(\mathbf{x,y};\theta )$ is a scoring function; $\mathbf{f}_{bs}(\mathbf{x},i,y_{i-1},y_i)$ is the feature vector at the $i^{th}$ character and $\theta$ is the feature weight vector. 
Please note that $c_0$ and $c_{n+1}$ are two pseudo characters marking the beginning and end of the sentence.
We use the features described in zhang et al. (2014) \cite{zhang-p14-character-parsing}, as shown in Table \ref{tab-template}.

\begin{table}[tb]
\caption{\label{tab-template} Feature templates for $\mathbf{f}_{bs}(\mathbf{x}, i, y_{i-1}, y_i)$ used in the baseline CRF model. 
$T(c_i)$ returns the type of the character $c_i$ (time, number, punctuation, special symbols, else).
$I(c_{i},c_{j})$ judges whether the two characters $c_i$ and $c_j$ are the same.
 }
\begin{center}
	\begin{tabular}{l c|l}
	\hline
\multicolumn{2}{c|}{Unigram: $\mathbf{f}_{bs\_uni}(\mathbf{x}, i, y_i)$}   &  \multicolumn{1}{c}{Bigram: $\mathbf{f}_{bs\_bi}(\mathbf{x}, i, y_{i-1}, y_i)$}    \\ \hline
01: $y_i \circ c_k$ & $i-2 \leq k \leq i+2 $  
	& 09: $y_{i-1} \circ y_i$ \\
02: $y_i \circ c_{k-1} \circ c_{k}$ &   $i-1 \leq k \leq i+2 $ 
	& 10: $y_{i-1} \circ y_i  \circ c_i $ \\
03: $y_i \circ c_{k-1} \circ c_{k} \circ c_{k+1}$ & $i-1 \leq k \leq i+1 $ 
	& 11: $y_{i-1} \circ y_i  \circ c_{i-1} \circ c_i$ \\
04: $y_i \circ T(c_k)$ & $i-1 \leq k \leq i+1 $  &\\
05: $y_i \circ T(c_{k-1}) \circ T(c_k)$ & $i \leq k \leq i+1$ & \\
06: $y_i \circ T(c_{i-1}) \circ T(c_i) \circ T(c_{i+1})$ &  &\\
07: $y_i \circ I(c_i,c_k)$ &  $i-2 \leq k \leq i+2, k \ne i$ & \\
08: $y_i \circ I(c_{i-1},c_{i+1})$ & & \\
\hline
	\end{tabular}
\end{center}
\end{table}

\section{Exploring External Lexicon Features}
\label{sec:lexicon}
Inspired by the work of Yu et al., (2015) \cite{YuZhengting-nlpcc2015-ws-pos} who have participated last year's shared task, 
we try to enhance the baseline CRF by using a large-scale word dictionary \cite{ZhangMeishan-jcip12}. 
The dictionary we use is composed of two parts.
The first part contains about $210K$ words, and is directly borrowed from Yu et al., (2015) \cite{YuZhengting-nlpcc2015-ws-pos}.\footnote{
  We are very grateful for their kind sharing. Their dictionary is composed of several word lists, the SogouW word dictionary (\url{http://www.sogou.com/labs/resource/w.php}), and a few lists on different domains (finance, sports, and entertainment) from the lexicon sharing website of Sogou (\url{http://pinyin.sogou.com/dict/}).
}
The second part contains $217K$ words, and is collected by ourselves from the lexicon sharing website of Sogou (\url{http://pinyin.sogou.com/dict/}).
In total, the external lexicon consists of $428,101$ words, and is denoted as 
 $\mathcal{D}$ in this work.

Apart from the features used in Table \ref{tab-template}, denoted as $\mathbf{f}_{bs}(\mathbf{x}, i, y_{i-1}, y_i)$, 
the enhanced model adds extra lexicon features to the feature vector, denoted as $\mathbf{f}_{lex}(\mathbf{x}, i, y_i, \mathcal{D})$.
Thus, the scoring function becomes:

\begin{equation}\label{score-lexicon}
  \begin{split}
    Score(\mathbf{x,y};\theta ) =& \sum_{1\leq i\leq n+1}\theta \cdot \begin{bmatrix}
        \mathbf{f}_{bs}(\mathbf{x},i,y_{i-1},y_i) \\
        \mathbf{f}_{lex}(\mathbf{x}, i, y_i, \mathcal{D})
        \end{bmatrix}
    \end{split}
\end{equation}
where the first term of the extended feature vector is the same as the baseline feature vector and 
the second term is the lexicon feature vector.

Table \ref{dictionary-template} lists the lexicon feature templates, which are mostly borrowed from Zhang et al. (2012) \cite{ZhangMeishan-jcip12}.
$F_B(\mathbf{x}, i, y_i, \mathcal{D})$ considers words beginning with $c_i$, and returns the maximum length $m$, so that the span $c_ic_{i+1}...c_{i+m-1}$ in $\mathbf{x}$ is a word in $\mathcal{D}$. ``Maximum'' means that there is no $r>m$ so that $c_ic_{i+1}...c_{i+r-1}$ in $\mathbf{x}$ is a word in $\mathcal{D}$.
In contrast, 
$F_E(\mathbf{x}, i, y_i, \mathcal{D})$ considers words ending with $c_i$, and returns the maximum length $m$, so that the span $c_{i-m+1}...c_{i-1}c_{i}$ in $\mathbf{x}$ is a word in $\mathcal{D}$. 
Analogously, 
$F_I(\mathbf{x}, i, y_i, \mathcal{D})$ considers words containing $c_i$ (absolutely inside), and returns the maximum length $m$, so that the span $c_{i-(m-j-1)}...c_{i}...c_{i+j}$ (where $m>2$ and $0<j<m-1$) in $\mathbf{x}$ is a word in $\mathcal{D}$. 

\begin{table}[tb]
\caption{\label{dictionary-template} Lexicon Feature templates $\mathbf{f}_{lex}(\mathbf{x}, i, y_i, \mathbf{D})$. }
\begin{center}
  \begin{tabular}{lll}
  \hline
01: $F_B(\mathbf{x}, i-1, y_{i}, \mathcal{D})$
  & 04: $F_B(\mathbf{x}, i, y_i, \mathcal{D})$
  & 07:  $F_B(\mathbf{x}, i+1, y_{i}, \mathcal{D})$ \\
02: $F_I(\mathbf{x}, i-1, y_{i}, \mathcal{D})$
  & 05: $F_I(\mathbf{x}, i, y_i, \mathcal{D})$
  & 08: $F_I(\mathbf{x}, i+1, y_{i}, \mathcal{D})$ \\
03: $F_E(\mathbf{x}, i-1, y_{i}, \mathcal{D})$ 
  & 06: $F_E(\mathbf{x}, i, y_i, \mathcal{D})$
  & 09: $F_E(\mathbf{x}, i+1, y_{i}, \mathcal{D})$ \\
\hline
  \end{tabular}
\end{center}
\end{table}

\section{The Guide-feature Based Approach for Exploiting \textsl{CTB7} and \textsl{PD}}
\label{sec:guide}

\begin{figure}[tb]
\begin{minipage}[t]{0.45\textwidth}
\centering
    \includegraphics[scale=0.7]{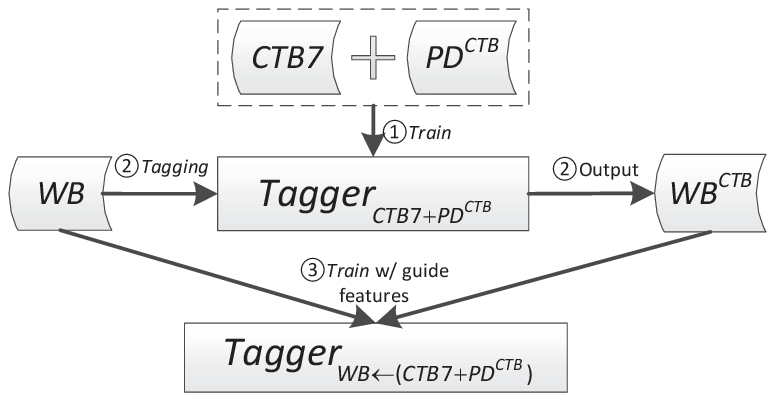}
\vspace{-1em}
  \caption{\label{fg-guide} Our model using guide feature}
\end{minipage}
\begin{minipage}[t]{0.55\textwidth}
\centering
    \includegraphics[scale=0.7]{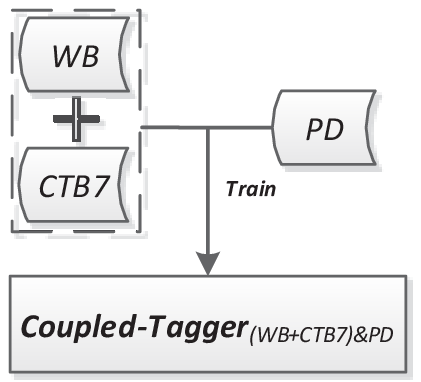}
\vspace{-1em}
  \caption{\label{fg-coupled-seg} Graphical structure of the coupled CRF}
\end{minipage}
\end{figure}

To use the heterogeneous data, we re-implement the guide feature baseline method \cite{jiang-p09}.
The basic idea is to use one resource to generate extra guide features on another resource, as illustrated in Fig. \ref{fg-guide}.
\textsl{PD} is converted into the style of \textsl{CTB}, as discussed in Section \ref{sec:exp-data}. 
First, we use $\textsl{CTB7}$ and $\textsl{PD}$ as the source data to train a source model $\textsl{Tagger}_{CTB7+PD}$. 
Then, $\textsl{Tagger}_{CTB7+PD}$ generates automatic tags on the target data $\textsl{WB}$, called $\textsl{source annotations}$. 
Finally, a target model $\textsl{Tagger}_{WB\leftarrow(CTB7+PD)}$ is trained on $\textsl{WB}$, using source annotations as extra guide features.

Table \ref{guide-template} lists the guide feature templates used in this work.  
Adding the guide features into the model feature vector, the scoring function becomes:
\begin{equation}\label{score-guide}
  \begin{split}
    Score(\mathbf{x,y};\theta ) =& \sum_{1\leq i\leq n+1}\theta \cdot \begin{bmatrix}
        \mathbf{f}_{bs}(\mathbf{x},i,y_{i-1},y_i) \\
        \mathbf{f}_{guide}(\mathbf{x}, \mathbf{y}^S, i, y_i)
        \end{bmatrix}
    \end{split}
\end{equation}

\begin{table}[tb]
\caption{\label{guide-template} Guide feature templates for $\mathbf{f}_{guide}(\mathbf{x}, \mathbf{y}^S, i, y_i)$. where 
$\mathbf{f}_{bs\_uni}(\mathbf{x}, i, y_i) \circ y_i^S$ means that each feature template in $\mathbf{f}_{bs\_uni}(\mathbf{x}, i, y_i)$ is concatenated with $y_i^S$ to produce a new feature template.}
\begin{center}
  \begin{tabular}{ll}
  \hline
\multicolumn{2}{c}{Guide Features: $\mathbf{f}_{guide}(\mathbf{x},  \mathbf{y}^S, i, y_i)$} \\ \hline
01: $\mathbf{f}_{bs\_uni}(\mathbf{x}, i, y_i) \circ y_i^S$
  & 05: $y_i \circ y_{i-1}^S \circ y_i^S$ \\
02: $y_i \circ y_i^S$
  & 06: $y_i \circ y_i^S \circ y_{i+1}^S$ \\
03: $y_i \circ y_{i+1}^S$
  & 07: $y_i \circ y_{i-1}^S \circ y_{i+1}^S$ \\
04: $y_i \circ y_{i-1}^S$
  & 08: $y_i \circ y_{i-1}^S \circ y_i^S \circ y_{i+1}^S$ \\
\hline
  \end{tabular}
\end{center}
\end{table}

\section{The Coupled Approach for Exploring \textsl{CTB7} and \textsl{PD}}
\label{sec:coupled}

The coupled sequence labeling approach is proposed in our earlier work Li et al. (2015) \cite{zhenghua-p15}, and aims to learn and predict two heterogeneous annotations simultaneously.
The key idea is to bundle two sets of tags together, and build a conditional random field (CRF) based tagging model in the enlarged space of bundled tags with the help of \emph{ambiguous labeling}. To train our model on two non-overlapping datasets that each has only one-side tags, we 
transform a one-side tag into a set of bundled tags by concatenating the tag with every possible tag at the missing side according to a predefined context-free tag-to-tag mapping function, thus producing ambiguous labeling as weak supervision. 
The bundled tag space contains $4 \times 4=16$ tags in our task of WS.
Please refer to Chao et al. (2015) \cite{chao-nlpcc-2015} for the detailed description of the coupled WS tagging model.

\section{The Merge-then-re-decode  Ensemble Approach}
\label{sec:ensemble}

In this section, we propose a merge-then-re-decode ensemble approach to combine the outputs of different base models, which is inspired by the work of Sagae and Lavie (2006) \cite{sagae-naacl06-reparsing}. 
First, given a sentence ${\mathbf{x} = c_1...c_n}$, 
the outputs of several base models are treated as votes of character-wise tags with equal weights. 
For example, if three models assign $B$ to the character $c_i$, and only one model assigns $S$ to it, then the scores of tagging $c_i$ as $\{B, I, E, S\}$ are $\{3, 0, 0, 1\}$ respectively. 
In such way, we can get all scores for all characters in $\mathbf{x}$.
Then, we find the highest-scoring tag sequence using the Viterbi algorithm.

To avoid that the re-decode procedure outputs a tag sequence containing illegal transitions ($B \rightarrow S$, $B \rightarrow B$, $I \rightarrow B$, $I \rightarrow S$, $E \rightarrow I$, $E \rightarrow E$, $S \rightarrow I$, $S \rightarrow E$), we make a slight modification to the standard Viterbi algorithm. 
The basic idea is to throw away illegal transitions from $c_{i-1}$ to $c_i$ when searching the best partial tag sequences for $c_{1}...c_i$. 
Concretely, if we are searching the best tag sequences for $c_{1}...c_i$ with $c_i$ tagged as $B$, we 
only considers the results that tag $c_{i-1}$ as $E$ or $S$ (but neither $B$ nor $I$).

\section{Experiments} 
\label{sec:exp}

\begin{table}[tb]
\caption{\label{data}Data statistics}
\begin{center}
\begin{scriptsize}
  \begin{tabular}{c|c|r|r|r}
  ~~Dataset~~ & ~~Partition~~ & ~~Sentences~~ & ~~Words ~~ & ~~Characters~~\\
  \hline
      & train & 20,135  & 421,166   & 688,734 \\
    WB  & dev & 2,052   & 43,697    & 73,244  \\
      & test  & 8,592   & ---     & 315,857 \\
  \hline
      & train & 46,572 & 1,039,774 & 1,682,485    \\
   CTB7 & dev & 2,079  & 59,955    &  100,316   \\
      & test  & 2,796  & 81,578    &   134,149  \\
  \hline
   PD & train & 106,157 & 1,752,502 & 2,911,489 \\
  \hline
  \end{tabular}
  \end{scriptsize}
\end{center}
\end{table}

\subsection{Datasets} 
\label{sec:exp-data}

Table \ref{data} shows the datasets used in this work.
``WB'', short for Weibo, refers to the labeled data provided by the NLPCC 2016 shared task organizer.
Actually, the organizer also provides a large set of unlabeled $\textsl{WB}$ text, which is not considered in this work.

We adopt \textsl{CTB7} as a third-party resource and follow the suggestion in the data description guideline for data split.

We also use \textsl{PD} as another labeled resource. Since \textsl{PD} and \textsl{CTB7} have different word segmentation and POS tagging standards, we used a converted version of $\textsl{PD}$ following the style of \textsl{CTB} for the sake of simplicity in this work.

\textbf{Annotation Conversion: $\textsl{PD}^\textsl{CTB}$}. We directly use the coupled WS\&POS tagging model trained on $\textsl{CTB5}$ and $\textsl{PD}$ in Li et al. (2016) \cite{zhenghua-d16} for data conversion.
As pointed in Li et al (2015) \cite{zhenghua-p15}, the coupled model can be naturally used for annotation conversion via constrained decoding with the \textsl{PD}-side tags being fixed. 
After conversion, if a sentence in \textsl{PD} contains a character with a very low marginal probability ($<0.8$), we throw away the sentence to guarantee the data quality.
Finally, we get the $100K \textsl{PD}$ dataset in the same style of $\textsl{CTB7}$, denoted as $\textsl{PD}^\textsl{CTB}$.

For \textbf{evaluation metrics}, we adopt character-level accuracy, and the standard Precision (P), Recall (R), and F1 score.

\textbf{Training with multiple training datasets}: 
For some models (such as $\textsl{WSTagger}_\textsl{CTB7+PD}$ and $\textsl{CoupledWSTagger}_\textsl{WB\&CTB7+PD}$), we use two or three training datasets simultaneously. 
To balance the contribution of different datasets, we adopt the simple corpus-weighting strategy proposed in Li et al. (2015) \cite{zhenghua-p15}.
Before each iteration, we randomly select $5000$ sentences from each training datasets.
Then, we merge and shuffle the selected sentences, and use them for one-iteration training.

\subsection{Heterogeneity of \textsl{WB} and \textsl{CTB7}} 
\label{sec:wb-vs-ctb7}

To investigate the heterogeneity of \textsl{WB} and \textsl{CTB7}, we use the baseline model trained on \textsl{WB}-train, denoted as $\textsl{WSTagger}_\textsl{WB}$, to process \textsl{CTB7}-dev/test, and also use the baseline model trained on \textsl{CTB7}-train, denoted as $\textsl{WSTagger}_\textsl{CTB7}$, to process \textsl{WB}-dev. Table \ref{ctb7-wb-compare} shows the results.
It is obvious that \textsl{CTB7} and \textsl{WB} differs a lot in the definition of word boundaries.
In contrast, in the shared task of NLPCC 2015, we find that \textsl{CTB7} and the provided \textsl{WB} are very similar in the word boundary standard \cite{chao-nlpcc-2015}.

Based on this observation, we employ the guide-feature based approach and the coupled approach to exploit \textsl{CTB7}, instead of directly adding \textsl{CTB7} as extra training data.

\setlength{\tabcolsep}{5pt}
 \begin{table*}[tb]
      \caption{\label{ctb7-wb-compare} WS accuracy: an investigation of the heterogeneity of \textsl{WB} and \textsl{CTB7}.} 
 \begin{center}
     \begin{scriptsize}
     \begin{tabular}{l | c  c |  c}    

 \multicolumn{1}{c|}{  } &  \multicolumn{2}{c|}{on \textsl{CTB7}} &  \multicolumn{1}{c}{on \textsl{WB}} \\
 \cline{2-4}
  & \multicolumn{1}{c}{dev} & \multicolumn{1}{c|}{{test}}
  & \multicolumn{1}{c}{dev}  \\ 
    \hline
        \multirow{1}{*}{$\textsl{WSTagger}_\textsl{CTB7}$ }
                & \textbf{96.37}
                & \textbf{95.81}
                  & 91.77 \\
        \multirow{1}{*}{$\textsl{WSTagger}_\textsl{WB}$}
                & 90.86
                & 90.82
                  & \textbf{94.66} \\
        \hline
     \end{tabular}
     \end{scriptsize}
   \end{center}
 \end{table*}

\subsection{Results on CTB7-dev/test} 
\label{sec:results-ctb7-dev-test}

\setlength{\tabcolsep}{1pt}
 \begin{table*}[t]
      \caption{\label{ctb7-result} Results on $\textsl{CTB7}$-dev/test.} 
 \begin{center}
     \begin{scriptsize}
     \begin{tabular}{l | l l l l | l l l l}    

 \multicolumn{1}{c|}{  } &  \multicolumn{4}{c|}{on Dev} &  \multicolumn{4}{c}{on Test} \\
 \cline{2-9}
 & \multicolumn{1}{c}{Acc}  & \multicolumn{1}{c}{P} & \multicolumn{1}{c}{\textsl{R}} & \multicolumn{1}{c|}{\textsl{F}} 
 & \multicolumn{1}{c}{Acc}  & \multicolumn{1}{c}{P} & \multicolumn{1}{c}{\textsl{R}} & \multicolumn{1}{c}{\textsl{F}} \\ 
    \hline
       \multirow{1}{*}{$\textsl{WSTagger}_\textsl{CTB7}$}
                & 96.37
                & 95.84
                & 95.37
                & 95.60 
                    & 95.81
                & 95.40
                & 94.58
                & 94.98  \\
        \multirow{1}{*}{$\textsl{WSTagger}_\textsl{CTB7+PD}$}
                & 96.82
                & 96.29
                & 96.14
                & 96.21 
                    & 96.37
                & 95.94
                & 95.44
                & 95.69 \\
      \hline
        \multirow{1}{*}{$\textsl{WS}$\&$\textsl{POSTagger}_\textsl{CTB7}$ }
                & 96.70
                & 96.21
                & 95.78
                & 96.00 
                    & 96.25
                & 95.92
                & 95.13
                & 95.52 \\
        \multirow{1}{*}{$\textsl{WS}$\&$\textsl{POSTagger}_\textsl{CTB7+PD}$}
                 & \textbf{97.04}
                 & \textbf{96.62}
                 & \textbf{96.34}
                 & \textbf{96.48}
                    & \textbf{96.61}
                 & \textbf{96.30}
                 & \textbf{95.66}
                 & \textbf{95.98}  \\

       \hline \hline
       \multirow{1}{*}{$\textsl{CoupledWSTagger}_\textsl{WB\&CTB7}$}
                & 96.54
                & 96.03
                & 95.55
                & 95.79 
                    & 96.02
                & 95.59
                & 94.86
                & 95.22 \\
       \multirow{1}{*}{$\textsl{CoupledWSTagger}_\textsl{WB\&CTB7+PD}$}
                & 96.96
                & 96.43
                & 96.21
                & 96.32 
                    & 96.45
                & 95.96
                & 95.48
                & 95.72\\
        \hline
        \multirow{1}{*}{$\textsl{CoupledWSTagger}_\textsl{WB\&CTB7 ~~}$ w/ lexicon}
                & {96.82}
                & {96.29}
                & {95.96}
                & {96.12}
                    & {96.42}
                & {95.95}
                & {95.39}
                & {95.67} \\
        \multirow{1}{*}{$\textsl{CoupledWSTagger}_\textsl{WB\&CTB7+PD~~}$ w/ lexicon} 
                & $\textbf{97.25}$
                & $\textbf{96.79}$
                & $\textbf{96.51}$
                & $\textbf{96.65}$
                    & $\textbf{96.83}$
                & $\textbf{96.45}$
                & $\textbf{95.88}$
                & $\textbf{96.16}$ \\
        \hline
     \end{tabular}
     \end{scriptsize}
   \end{center}
 \end{table*}

\setlength{\tabcolsep}{1pt}
 \begin{table*}[t]
      \caption{\label{ctb7-pos-result} Performance of joint WS\&POS tagging on $\textsl{CTB7}$-dev/test.} 
 \begin{center}
     \begin{scriptsize}
     \begin{tabular}{l | l l l |  l l l}    

 \multicolumn{1}{c|}{  } &  \multicolumn{3}{c|}{on Dev} &  \multicolumn{3}{c}{on Test} \\
 \cline{2-7}
  & \multicolumn{1}{c}{P} & \multicolumn{1}{c}{\textsl{R}} & \multicolumn{1}{c|}{\textsl{F}} 
  & \multicolumn{1}{c}{P} & \multicolumn{1}{c}{\textsl{R}} & \multicolumn{1}{c}{\textsl{F}} \\ 
    \hline
        \multirow{1}{*}{$\textsl{WS}$\&$\textsl{POSTagger}_\textsl{CTB7}$ }
                & 91.28
                & 90.86
                & 91.04 
                  & 90.91
                  & 90.16
                  & 90.54 \\
        \multirow{1}{*}{$\textsl{WS}$\&$\textsl{POSTagger}_\textsl{CTB7+PD}$}
                 & \textbf{92.19}
                 & \textbf{91.92}
                 & \textbf{92.06}
                    & \textbf{91.80}
                    & \textbf{91.19}
                    & \textbf{91.49}  \\
        \hline
     \end{tabular}
     \end{scriptsize}
   \end{center}
 \end{table*}

To investigate the performance on canonical texts of the models trained on \textsl{CTB7} (and \textsl{PD}), 
we evaluate the models on \textsl{CTB7}-dev/test.
Table \ref{ctb7-result} shows the results on the task of WS.
We can get several reasonable yet interesting findings.
First, comparing the results in all four major rows, we can see that using $\textsl{PD}$ as extra labeled data consistently improves the F1 score by about $0.5\%$.
Second, comparing the results in the first two major rows, it is clear that jointly modeling WS\&POS outperforms the pure WS tagging model by about $0.3-0.5\%$. 
Third, comparing the results in the bottom two major rows, we can see that lexicon features are useful and improves F1 score by about $0.5\%$.
Fourth, comparing the results in the first and third major rows, we can see that using WB as extra labeled data leads with the coupled approach to slight improvement in F1 score ($0.03-0.24\%$).

Table \ref{ctb7-pos-result} shows the results on the joint task of WS\&POS.
We can see that using \textsl{PD} as extra labeled data dramatically improves the word-wise F1 score by about $1\%$.

\subsection{Results on WB-dev} 
\label{sec:exp-ws}

In this part, we conduct extensive experiments to investigate the effectiveness of different methods for WS on \textsl{WB}-dev. Table \ref{seg-result-ultimate} shows the results. From the results, we can obtain the following findings.


First, lexicon features are very useful. Comparing the first two major rows, we can see that using lexicon features leads to a large improvement of $94.88-93.65=1.23\%$ on F1 score over the baseline model.
Comparing the third and fourth major rows, lexicon features boost F1 score by $95.15-94.16=0.99\%$ over the models with guide features.
Comparing the fifth and sixth major rows, lexicon features boost F1 score by $95.30-94.64=0.66\%$ over the coupled models.

Second, the coupled approach is much more effective than the guide-feature based approach in exploiting multiple heterogeneous data. 
Comparing the third and fifth major rows, the coupled approach outperforms the guide-feature based approach by $94.64-94.16=0.48\%$ on F1 score. 
Comparing the fourth and sixth major rows, with the lexicon features, the coupled approach achieves higher F1 score by $95.30-95.15=0.15\%$ over its counterpart.

Third, looking into the third major row, we also get a few interesting findings: 
1) using a joint WS\&POS tagger to produce guide tags is better than using a WS tagger, indicating that jointly modeling WS\&POS leads to better guide information, which is consistent with the results in Table \ref{ctb7-result}; 
2) \textsl{PD} is helpful by producing better guide tags, leading to higher F1 score on \textsl{WB}-dev by about $0.2\%$;
3) using both WS\&POS tags for guide achieves nearly the same performance as using only WS tags.

Finally, the proposed merge-then-re-decode ensemble approach improves F1 score by $95.47-95.30=0.17\%$ over the best single model.
However, we find that the performance drops when we use all model during ensemble, which may be caused by the very bad performance of some models.

\setlength{\tabcolsep}{1pt}
 \begin{table*}[t]
      \caption{\label{seg-result-ultimate}Results on $\textsl{WB}$-dev} 
 \begin{center}
     \begin{scriptsize}
     \begin{tabular}{c l | l l l l}    

 \multicolumn{2}{c|}{ Approaches }
 & \multicolumn{1}{c}{Acc}  & \multicolumn{1}{c}{P} & \multicolumn{1}{c}{\textsl{R}} & \multicolumn{1}{c}{\textsl{F}} \\ 
 		\hline
       \multirow{1}{*}{Baseline}  &  1.$\textsl{WSTagger}_\textsl{WB}$ & 94.66 
       					   & 93.30
 				           & 93.99 
 				           & 93.65 \\
       \hline
        \multirow{1}{*}{w/ lexicon features} & 2.$\textsl{WSTagger}_{\textsl{WB}}$ & 95.74
 			        & 94.45
 			        & 95.31
 			        & 94.88 \\
       \hline
       \multirow{6}{*}{w/ guide features } 
              		& 3.WS-tag from $\textsl{WSTagger}_\textsl{CTB7}$ & 94.52
       					& 93.21
       					& 93.93
       					& 93.58 \\
       				& 4.WS-tag from $\textsl{WSTagger}_\textsl{CTB7+PD}$ & 94.80
       					& 93.41
       					& 94.40
       					& 93.90 \\
       				& 5.WS-tag from $\textsl{WS}$\&$\textsl{POSTagger}_\textsl{CTB7}$ & 94.86
       					& 93.64
       					& 94.27
       					& 93.95 \\
       				& 6.WS-tag from $\textsl{WS}$\&$\textsl{POSTagger}_\textsl{CTB7+PD}$ & \textbf{95.05}
       				 	 & 93.76
       				 	 & \textbf{94.57}
       				 	 & \textbf{94.16} \\
       				& 7.WS\&POS-tag from $\textsl{WS}$\&$\textsl{POSTagger}_\textsl{CTB7}$ & 94.88
       					& \textbf{94.33}
       					& 93.64
       					& 93.98 \\
       				& 8.WS\&POS-tag from $\textsl{WS}$\&$\textsl{POSTagger}_\textsl{CTB7+PD}$ & 95.03
      					 & 93.83
      					 & 94.50
       					 & \textbf{94.16} \\
       	\hline
       	\multirow{1}{*}{w/ lexicon \& guide} & 9.WS\&POS-tag from $\textsl{WS\&POSTagger}_\textsl{CTB7+PD}$ & 95.97
       					& 94.77
       					& 95.53
       					& 95.15 \\
       \hline
       \multirow{2}{*}{Coupled} & 10.$\textsl{CoupledWSTagger}_\textsl{WB\&CTB7}$ & 95.38
       					& 94.12
       					& 94.91
       					& 94.51 \\
       				& 11.$\textsl{CoupledWSTagger}_\textsl{WB\&CTB7+PD}$ & \textbf{95.50}
       					& \textbf{94.25}
       					& \textbf{95.03}
       					& \textbf{94.64} \\
       	\hline
       	\multirow{3}{*}{Coupled  w/ lexicon} &	 12.$\textsl{CoupledWSTagger}_\textsl{WB\&CTB7}$ & 96.01
       					& 94.74
       					& 95.61
       					& 95.17 \\
       				& \multirow{1}{*}{13.$\textsl{CoupledWSTagger}_\textsl{WB\&CTB7+PD}$} (submitted) & 95.98 
       					& 94.78
       					& 95.56
       					& 95.17 \\
       				& \multirow{1}{*}{14.$\textsl{CoupledWSTagger}_\textsl{WB\&CTB7+PD}$} & \textbf{96.11}
       					& \textbf{94.80}
       					& \textbf{95.82}
       					& \textbf{95.30} \\
       	\hline
       	\multirow{3}{*}{Merge-then-re-decode} & On four models (2,9,12,13) (submitted) & 96.14
       					& 95.03
       					& 95.72
       					& 95.37 \\
       				& On four models (2,9,12,14) & $\mathbf{96.22}$
       					& $\mathbf{95.10}$
       					& $\mathbf{95.84}$
       					& $\mathbf{95.47}$ \\
       				& On all models (w/o 13) & 95.88
                & 94.76
                & 95.48
                & 95.12 \\
       	\hline
     \end{tabular}
     \end{scriptsize}
   \end{center}
 \end{table*}

\subsection{Reported results on WB-test}
\label{sec-final}

Since we do not have the gold-standard labels for the test data, Table \ref{fin-result}
 shows the results provided by the shared task organizers. 
Our effort leads to an improvement on WS F1 score by $95.37-93.83=1.54\%$. 
And our results on test data rank the first place among five participants, and is also the best among all 28 submissions. 

\setlength{\tabcolsep}{5pt}
\begin{table}[t]
\caption{\label{fin-result} Results on \textsl{WB}-test}
\begin{center}
\begin{scriptsize}
	\begin{tabular}{l|l|l|l}
	\multicolumn{1}{c|}{} & \multicolumn{1}{l|}{P}&  \multicolumn{1}{l|}{R} & \multicolumn{1}{l}{F} \\
	\hline
	Baseline & 93.53&  94.14 & 93.83 \\
	\hline
	Merge-then-re-decode (2,9.12,13) & 95.05 (\textbf{+1.52})&  95.70 (\textbf{+1.56})  & 95.37 (\textbf{+1.54}) \\	
	\hline
	\end{tabular}
\end{scriptsize}
\end{center}

\end{table}

\section{Related Work}
\label{sec:related}

Using external lexicon is first described in Pi-Chuan Chang et al. (2008) \cite{chang-2008-word-seg-lexicon}.
Zhang et al. (2012) \cite{ZhangMeishan-jcip12} find the lexicon features are also very helpful for domain adaptation of WS models, 

Jiang et al. (2009) \cite{jiang-p09} first propose the simple yet effective guild-feature based method, which is further extended in \cite{SunWeiwei-p11,zhenghua-p12-multi-treebank,sun-p12-heterogeneous}.

Qiu et al. (2013) \cite{qiu-d13-heterogeneous} propose a model that performs heterogeneous Chinese word segmentation and POS tagging and produces 
two sets of results following $\textsl{CTB}$ and $\textsl{PD}$ styles respectively.
Their model is based on linear perceptron, and uses approximate inference.

Li et al. (2015) \cite{zhenghua-p15} first propose the coupled sequence labeling approach.
Chao et al., (2015) \cite{chao-nlpcc-2015} make extensive use of the coupled approach in participating the NLPCC 2015 shared task of WS\&POS for Webo texts.
Li et al., (2016) \cite{zhenghua-d16} further improves the coupled approach in terms of efficiency via context-aware pruning, and first apply the coupled approach to the joint WS\&POS task. 
In this work, we directly use the coupled model built in Li et al. for converting the WS\&POS annotations in \textsl{PD} into the style of \textsl{CTB}.

\section{Conclusion}
\label{sec:conclude}
We have participated in the NLPCC 2016 shared task on Chinese WS for Weibo Text. 
Our main focus is to make full use of an external lexicon and two heterogeneous labeled data (i.e., \textsl{CTB7} and \textsl{PD}).
Moreover, we apply an merge-then-re-decode ensemble approach to combine the outputs of different base models. 
Extensive experiments are conducted in this work to fully investigate the effectiveness of methods in study. 
Particularly, this work leads to several interesting findings.
First, lexicon features are very useful in improving performance on both canonical texts and WB texts.
Second, the coupled approach is consistently more effective than the guide-feature based approach in exploiting multiple heterogeneous data.
Third, using the same training data, a joint WS\&POS model produces better WS results than a pure WS model, indicating that the POS tags are helpful for determining word boundaries.
Our submitted results rank the first place among five participants in the open track in terms of F1 score, and is also the best among all 28 submissions.

For future work, we plan to work on word segmentation with different granularity levels. 
During this work, we carefully compared the outputs of different base models, and found that in many error cases, the results of the statistical models are actually correct from the human point view. 
Many results are considered as wrong answers simply because they are of different  word granularity from the gold-standard references.
Therefore, we are very interested in build statistical models that can output WS results with different granularities. And perhaps, we have to first construct some WS data with multiple-granularity annotations.

\section*{Acknowledgments}
The authors would like to thank the anonymous reviewers for the helpful comments.
This work was supported by National Natural Science Foundation of China (Grant No. 61502325, 61432013) and the Natural Science Foundation of the Jiangsu Higher Education Institutions of China (No. 15KJB520031).  



\bibliographystyle{splncs03}
\bibliography{reference}

\end{document}